\documentclass[letterpaper]{article} 
\usepackage{aaai2026}  
\nocopyright  
\usepackage{times}  
\usepackage{helvet}  
\usepackage{courier}  
\usepackage[hyphens]{url}  
\usepackage{graphicx} 
\usepackage{natbib}  
\usepackage{caption} 
\usepackage{booktabs}
\usepackage{amsmath}
\usepackage{amssymb}
\usepackage{xcolor}
\definecolor{linknavy}{RGB}{18,66,122}
\usepackage[colorlinks=true,linkcolor=linknavy,citecolor=linknavy,urlcolor=linknavy,
            pdfborder={0 0 0}]{hyperref}
\makeatletter\expandafter\let\csname ver@hyperref.sty\endcsname\relax\makeatother
\graphicspath{{figures/}}

\newcommand{\helps}{\textbf{helps}}
\newcommand{\nofx}{\textbf{no effect}}
\newcommand{\fails}{\textbf{fails}}
\newcommand\blfootnote[1]{\begingroup\renewcommand\thefootnote{}\footnotetext{#1}\endgroup}

\title{A Gold-Standard Study of What Makes\\
a Lightweight Game-Playing Agent Strong}
\author{
    Nima Kelidari\equalcontrib,
    Mohammadsaeed Haghi\equalcontrib,
    Mahdi Salmani\equalcontrib
}
\affiliations{
    University of Southern California\\
    \{kelidari, haghim, salmanis\}@usc.edu
}

\newcommand{\ORACLEOBSMEAN}{24.7}
\newcommand{\ORACLEOBSN}{four}
\begin{document}
\maketitle

\begin{center}\small
Accepted as a poster at the 22nd AAAI Conference on Artificial Intelligence and
Interactive Digital Entertainment (AIIDE 2026).
\end{center}
\blfootnote{Code, trained models, and the human-play client: \url{https://github.com/Nikelroid/adversarial-coevolution}.}

\begin{abstract}
\noindent
A reinforcement learning agent is only as strong as the opponents it trains
against, and it is also hard to grade. It beats a random opponent over 99 percent of the time,
and it only ties copies of itself. So we build a strong, fixed, rule-based expert for Gin Rummy
and never train against it: no expert games, no demonstrations, no gradients. It does pick the
shipped checkpoint, and we say where that matters. The expert beats every agent we trained 70 to
99 percent of the time. Across more than a hundred runs we ask what makes a small agent
stronger, and five things help. They are trust-region updates, a well-aimed reward, a curriculum
of tougher opponents, warm starting, and keeping the best checkpoint. Stacking them lifts a
self-play champion from about 30 to about 34 percent against the expert. Several other ideas did
not pay off. Reward shaping at two horizons, learned state embeddings, and DAgger imitation
were each unhelpful, and a live large language model opponent was far too slow. Extra capacity
does little to break the ceiling, across MLP, convolutional, set-based, attention and recurrent
encoders. Handing the learner the opponent's hidden hand does not move it either. We add two standard baselines: neural fictitious self-play and information-set
Monte Carlo search. We also reproduce the grading protocol on Leduc Hold'em, whose optimum is
computable. The result is a
small, fast recipe that never trains on the expert, plus a grading protocol that carries over to
any game with a fixed reference. We report bootstrap intervals on the multi-seed comparisons and
release the code.
\end{abstract}

\section{Introduction}
Imperfect-information card games are a long-standing testbed for game AI.
Poker, DouDizhu, Mahjong, and Gin Rummy all hide the opponent's hand and the order of the deck.
A player must act under uncertainty, plan over a long horizon, and adapt to whoever sits across
the table. The usual recipe is deep reinforcement learning (RL) with self-play, and it has
produced expert-level play in several of these games \citep{nfsp,suphx,douzero}. Behind those
results, a practitioner who wants a small, fast agent hits two problems.

The first is the \emph{opponent bottleneck}. An agent is only as good as the
opponents it practices against, and a weak fixed opponent teaches a weak ceiling. Pure self-play
can chase its own tail, cycling through strategies that beat the last version without getting
any stronger. The second problem is that most of these games offer \emph{no cheap yardstick}.
Strength is usually measured against random play or against the agent's own past checkpoints,
and both can flatter a mediocre agent. So the choices that build a strong small agent get tuned
against a reference that is too weak or always moving. Those choices are the algorithm, the
reward, the opponent schedule, the state representation, and which checkpoint to ship. The field
has many strong single agents, but few controlled accounts of which choices matter.

Existing work does not fill this gap. Large self-play systems
\citep{alphazero,alphastar,openaifive} answer what very large compute can do. Equilibrium
methods \citep{cfr,libratus,deepcfr} play extremely well, but they do not prescribe a reward or
a curriculum for a small reactive policy. And prior Gin Rummy work studies single methods in
isolation \citep{tdrummy,ginensemble} rather than comparing them against one fixed
reference.

So we build a strong, fixed, deterministic expert for Gin Rummy. It solves the
one subproblem with a clean optimum, meld decomposition, which is the lowest-deadwood way to
arrange a hand. Every other move follows a fixed rule. The result is a strong, cheap, reproducible heuristic rather than a
game-theoretic optimum for the full game. A fair yardstick here needs a fixed heuristic, not
an optimum. We never train against it. We then run more than a hundred
controlled experiments, each changing one factor and each graded against the expert. Our subject
is the study, not the expert.

\begin{itemize}
\itemsep0.15em
\item A reproducible expert benchmark for Gin Rummy, and the finding that strong
play almost never gins: the expert gins in under two percent of games. We built it to knock early, so that part is an assumption. But an ablation that
holds out for gin drops its win-rate from $70$ to $15$ percent. And no reward moved a learner
the other way: paying three times more for a gin leaves the gin rate under one percent.
\item A controlled study, graded against the fixed expert, of which lightweight training
choices help. Five choices help: trust-region updates, a knock-first reward, a rising opponent
curriculum, warm-starting, and keeping the best checkpoint. Four fail: learned state
embeddings, dense step rewards, imitation from a large-language-model (LLM) guide, and a live
LLM opponent.
\item No sign that capacity is what binds. MLP, convolutional, set, and recurrent encoders land
in one band with overlapping intervals across recipes, and attention matches a plain MLP under
PPO. A determinized search graded \emph{fairly} loses to our trained agents, while an
\emph{oracle} variant reading the hidden cards reaches far higher. The gap bounds what the
information is worth without pricing it, yet handing the same cards to the learner as an input
produces no gain.
\item A release of the pipeline. We reproduce the grading protocol on Leduc Hold'em,
where a counterfactual-regret optimum is computable and a tabular learner reaches near parity.
\end{itemize}

\section{Related Work}

\paragraph{Deep RL for imperfect-information card games.}
Self-play deep RL has mastered several hidden-information card and tile games.
Neural Fictitious Self-Play mixes a best-response network with an average-policy network
\citep{nfsp}, and Suphx reached expert human level at Mahjong \citep{suphx}. DouZero mastered
DouDizhu with parallel Monte-Carlo self-play \citep{douzero}, and RLCard supplies the common
toolkit and environments, Gin Rummy among them \citep{rlcard}. Within Gin Rummy,
\citet{tdrummy} compared temporal-difference self-play with coevolution, and the EAAI challenge
produced hand-tuned and ensemble agents \citep{ginensemble}. None of this work gives a controlled account of which training choices matter.
We address that gap by holding one expert fixed while measuring those choices.

\paragraph{Self-play, leagues, and opponent curricula.}
AlphaGo Zero and AlphaZero showed that pure self-play with search can reach
superhuman play in perfect-information games \citep{alphagozero,alphazero}. AlphaStar added a
league with prioritized fictitious self-play (PFSP), which weights opponents by the learner's
current win-rate against them \citep{alphastar}. OpenAI Five scaled self-play to a complex video
game \citep{openaifive}. Curriculum learning orders training from easy to hard
\citep{curriculum}, and a large literature studies curricula for RL \citep{curriculumsurvey}. We adopt a small opponent curriculum with a PFSP variant, then ask which
curriculum and checkpoint choices matter at small scale against a fixed expert.

\paragraph{Search and equilibrium solving.}
Counterfactual regret minimization \citep{cfr} and its deep variants
\citep{deepcfr} underpin the superhuman poker agents Libratus and Pluribus
\citep{libratus,pluribus}. OpenSpiel collects many such algorithms and small benchmark games
\citep{openspiel}. Determinized search samples the hidden state, then searches the resulting
perfect-information game, and information-set Monte-Carlo tree search (ISMCTS) is the canonical
form \citep{ismcts,pimc}. We use a determinized ISMCTS as the Gin Rummy \emph{baseline}. On the second
game, CFR provides the reference and NFSP the learned baseline. Later we show that a fair
determinized search is far weaker than the same search with oracle access.

\paragraph{Network architectures for card observations.}
A hand is an unordered set of cards, which invites permutation-invariant encoders
such as Deep Sets \citep{deepsets} and self-attention \citep{transformer}. Convolutional and
recurrent networks \citep{lstm} are the other standard choices. We hold the training recipe fixed and sweep these encoders. This tests whether
the network, rather than the reward or the curriculum, can break the ceiling.

\paragraph{Reward shaping and imitation learning.}
Potential-based reward shaping changes the learning signal while preserving the
optimal policy \citep{shaping}. We use that view to read why some shaped rewards help and
others hurt. DAgger reduces imitation learning to no-regret online learning, by querying a
teacher on the states the student visits \citep{dagger}. Imitation collapses here. One reading
is causal confusion, where a cloned policy latches onto spurious correlates of the teacher's
action and breaks under distribution shift \citep{causalconfusion}. Another is plainer: a
feed-forward policy may simply lack the structure to approximate the teacher's decision rule.
Our data do not separate the two.

\paragraph{Language models as game players.}
Cicero combined a language model with planning and RL to reach human-level play
in Diplomacy \citep{cicero}. We test an instruction-tuned LLM \citep{qwen} as a live opponent
inside the training loop. It is far too slow for the tens of thousands of opponent queries one
run issues.

\paragraph{Action masking and tooling.}
We use invalid-action masking to keep a policy-gradient agent inside the legal
action set \citep{mask}, with PettingZoo providing the multi-agent interface
\citep{pettingzoo}. PPO and
TRPO come from Stable-Baselines3 and its contrib package \citep{sb3,ppo,trpo,gae}. LLM
inference runs on a paged-attention server \citep{vllm}.

\section{Problem Setup}
\begin{figure*}[t]
\centering
\includegraphics[width=0.82\textwidth]{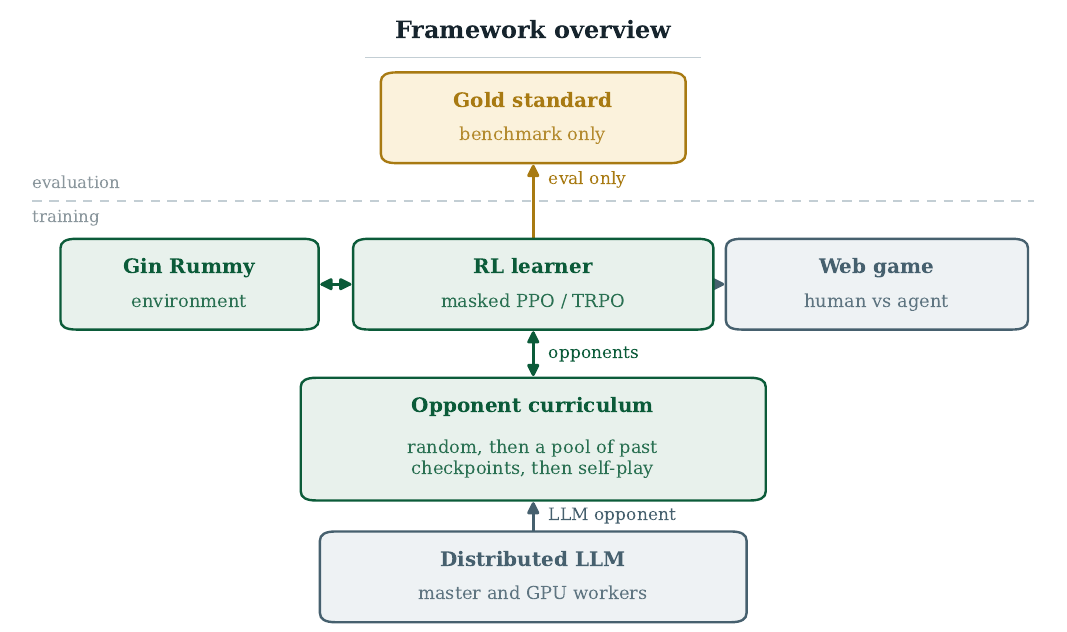}
\caption{System overview. A masked PPO or TRPO learner trains against a rising
curriculum of opponents (random play, a growing pool of past checkpoints, then
self-play), with a distributed LLM as an optional opponent. The fixed expert
sits outside the training loop and only grades agents.}
\label{fig:arch}
\end{figure*}
We study two-player Gin Rummy as implemented in RLCard and exposed through
PettingZoo as \texttt{gin\_rummy\_v4} \citep{rlcard,pettingzoo}; Figure~\ref{fig:arch} shows
how the pieces fit together.

\paragraph{The game.}
Gin Rummy is a two-player draw-and-discard game played with a standard $52$-card
deck, and each player holds ten cards. Cards group into \emph{runs} of three or more in a suit,
and \emph{sets} of three or four of a kind. No card serves two melds at once, and whatever is
left over is \emph{deadwood}. A turn draws the top of the stock or the discard pile, then
discards one card face up. A player may \emph{knock} once its deadwood is at most ten, or
declare \emph{gin} at zero deadwood. The knocker scores the opponent's deadwood minus its own,
and gin earns a fixed bonus. A defender whose deadwood is no larger \emph{undercuts} and scores
instead. Gin pays the most but demands zero deadwood, so the chance to knock arrives far more
often than the chance to gin.

\paragraph{Observation and actions.}
The agent sees a $4\times52$ binary tensor whose planes show its own hand, the
top discard, the other visible discards, and the unknown cards. Unknown cards are those in the
opponent's hand or the stock, and the agent never sees the opponent's hand. The action space
has $110$ discrete actions: draw, pick up, declare a dead hand, gin, two scoring actions, and one
discard or knock per card. At every step the environment supplies a binary mask of the legal
actions.

\paragraph{Single-agent reduction.}
We turn the two-player game into a single-agent problem. One seat is the learner
and the other is an opponent drawn from a curriculum. The wrapper steps that opponent to
completion between the learner's decisions, and it rotates seats so neither position is
favored. We can then swap in any opponent without touching the learner.

\paragraph{Masked actor-critic.}
PPO and TRPO share one masked actor-critic \citep{ppo,trpo}. Before the softmax
we set the logits of illegal actions to a large finite negative value. Using a finite value
rather than negative infinity is deliberate. PPO tolerates infinities, but TRPO's
conjugate-gradient and KL computations produce NaNs on them, so a finite fill keeps one network
usable by both. At evaluation we always take the highest-probability \emph{legal} action. We
never fall back to a random legal move, which would hide the states where the policy is
undecided.

\section{The Gold-Standard Expert}
The center of our method is a fixed reference opponent that is strong, cheap,
and fully reproducible. This section gives the design and the case for it; the
full rule-by-rule specification is in the released code.

\paragraph{Design.}
Given any hand, the expert computes the minimum achievable deadwood exactly by
enumerating every valid arrangement of runs and sets. The result is memoized, so the benchmark
stays fast. This is the one subproblem where the expert is provably optimal, and three fixed
rules do the rest. It draws the up-card only when that strictly lowers the best achievable
deadwood. It discards the card that minimizes the resulting deadwood, shedding the highest-value
card on ties. And it knocks as soon as the rules allow, taking gin only when gin costs no delay.
It never models the opponent's hidden cards, so it is a strong heuristic rather than a
game-theoretic optimum.

\paragraph{Why this is the right yardstick.}
A controlled study needs a reference that separates good agents from weak ones,
stays identical across comparisons, and grades thousands of games cheaply. Ours plays $600$
games in about two minutes.
We call it a \emph{gold standard} in that benchmarking sense, not in the sense of
game-theoretic optimality. It never appears in the training loop. No agent below practices against the
thing we measure with, and the imitation probe takes its demonstrations from a separate LLM
guide. It does grade checkpoints during training, as the next section records. So it selects the
shipped model, but it never trains it.

\paragraph{How strong, and why.}
Two measurements make the case. First, the expert is hard to beat even for a
search that cheats. An oracle determinized search may read the opponent's hidden cards. It still
needs $120$ rollouts per move to beat the expert $85$ percent of the time, and manages only $41$
percent at $10$ rollouts. The same search graded fairly never gets past $26$ percent
(Figure~\ref{fig:archbase}b). Second, the knock-early style we stipulated survives an ablation.
Switching the same agent to hold out for gin raises its gin rate to $96$ percent against random
play over $600$ games. Its win-rate against the self-play champion drops from $70$ to $15$
percent.

\section{Training Methods Studied}
All learning agents share the masked actor-critic above, and we vary one
ingredient at a time around it.

\paragraph{Algorithm.}
We compare PPO, which clips the policy update, against TRPO, which bounds it by a
trust region on the policy's KL divergence \citep{ppo,trpo}. Both use generalized advantage
estimation \citep{gae}, and both keep the same network, curriculum and reward. The algorithm is
the intended changed factor.

\paragraph{Reward shaping.}
The native reward is sparse, giving one score at the end of the game. We study
shaped rewards that change the relative payoff of knocking and ginning, plus a small dense
bonus for reducing one's own deadwood. Potential-based shaping \citep{shaping} is the framing we
reason with, as intuition rather than guarantee. Our deadwood bonus is not the difference of a
potential at consecutive states. So the policy-invariance result does not cover it, and any
shift it causes is a real change of objective.

\paragraph{Opponent curriculum.}
We schedule opponents in three stages: random play, then a growing pool of past
checkpoints, then a mix that adds self-play. The pool is sampled uniformly over recent
checkpoints. It can also use a PFSP weighting, which practices more against the opponents the
agent is losing to \citep{alphastar}. Stage boundaries are fractions of the training budget,
shared across workers through one state file.

\paragraph{Warm-start and keep-the-best.}
A run can warm-start from a strong earlier agent rather than from scratch.
Because shifting self-play drifts past its own peak, we evaluate at fixed
intervals and keep a copy whenever the agent improves. We ship that checkpoint
rather than the final one. Checkpoints are ranked by win-rate against the expert, with ties
broken against the champion. The expert therefore selects the checkpoint, though it never
supplies a gradient. Headline numbers re-measure that checkpoint over $2000$ games, and they
carry the optimism of a chosen maximum.

\paragraph{State representation.}
We test whether a compact dense embedding of the sparse $4\times52$ tensor helps,
and we build one two ways. The first is learned to place states with similar outcomes near each
other. For the second, an LLM judges state similarity and we fit an embedding to those
judgments. We try several sizes, both frozen and unfrozen.

\paragraph{Imitation and dense supervision.}
We test imitation from a strong but imperfect guide. The student plays, and we
query an LLM on the states the student reaches and clone its answers. We then continue with RL
to try to surpass it, which makes this a DAgger-style loop \citep{dagger}. The expert is not the
teacher here; it only scores the result. We also test a short-horizon dense reward that scores
each move by agreement with an evaluator over the next $L$ steps.

\paragraph{Live LLM opponent.}
We test whether an instruction-tuned LLM can serve as an in-the-loop opponent.
One run issues millions of environment steps and tens of thousands of opponent queries. So we
serve the model through a caching CPU master over a pool of GPU workers \citep{vllm,qwen}.

\paragraph{Network architecture.}
Holding every other setting fixed, we vary \emph{only} the policy network. Each
one trains from scratch, since a different-shaped network cannot warm-start from the MLP
champion. This is a relative ranking at a fixed budget. We sweep MLP width and depth, a convolutional stack over the card planes, and a
Deep Sets encoder \citep{deepsets}. We also sweep a self-attention encoder over card tokens
\citep{transformer}, an LSTM \citep{lstm}, activation functions and weight decay. We then rerun the top encoders under two other recipes, PPO in place of TRPO and
PFSP sampling, to see whether any gain is recipe-specific.

\paragraph{Search and learning baselines.}
We add two reference points from outside our recipe. The first is a determinized search with an information-set tree (ISMCTS)
\citep{ismcts,pimc}. Before each rollout it re-deals the cards it cannot see, the opponent's
hand and the stock, uniformly from the unseen pool. It never uses the true hidden cards, and we vary its rollout budget per move. For
contrast, an \emph{oracle} variant searches from the true state as an upper bound. The second is Neural
Fictitious Self-Play \citep{nfsp}, the canonical equilibrium-learning baseline,
run through OpenSpiel \citep{openspiel}.

\paragraph{A second game.}
We also check that the protocol is not tied to Gin Rummy. Leduc Hold'em is a small
poker game where counterfactual regret minimization gives a computable near-optimal expert
\citep{cfr,openspiel}. A tabular self-play learner and NFSP are graded against it, exactly as
our Gin Rummy agents are graded against the fixed expert.

\section{Experimental Setup}
\paragraph{One metric, applied the same way everywhere.}
Every agent is graded by win-rate against the fixed expert, with seats rotated so
each side plays both positions equally. Using one fixed metric isolates the ingredient we
changed. Where
useful we add win-rate against the prior self-play champion and against random play. Headline
agents get $2000$ games with a $95$ percent confidence interval. Broad sweeps get $400$ to $600$
games per cell, enough to rank regimes but not to separate neighbours. For multi-seed
comparisons we report the interquartile mean (IQM) with stratified bootstrap intervals, not a
single best run \citep{rliable,deeprlmatters}.

\paragraph{Protocol.}
Each study changes one factor and holds the rest fixed. Algorithm runs use two random seeds at two million steps. The reward and
curriculum sweep changes one setting at a time from a common baseline. Representation runs swap only the input. A win-rate counts a
positive game score as a win.

\section{Results}

\begin{figure*}[t]
\centering
\includegraphics[width=0.94\textwidth]{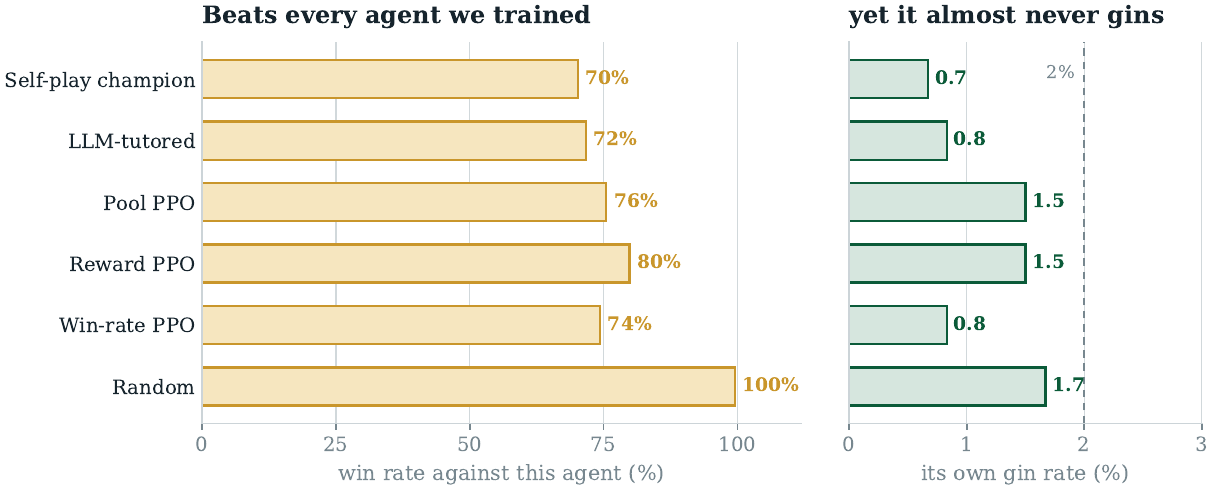}
\caption{The fixed expert beats every agent we trained (left) yet gins in under
two percent of games (right). It wins by knocking early with low deadwood, not
by chasing gin.}
\label{fig:gold}
\end{figure*}

\subsection{The expert is strong, and it almost never gins}
The expert beats every agent we trained between $70$ and $99$ percent of the
time. It takes $70.2$ percent against the self-play champion and $99.5$ percent against random
play (Figure~\ref{fig:gold}, left). The surprise is on the right of the same figure. Gin scores
the most points, yet the expert gins in only $0.7$ to $1.7$ percent of games. It wins while
almost never ginning.

We built the expert to knock as soon as it legally can. Its win rate therefore
cannot show, on its own, that knocking early is right. The strategy was an assumption we made,
not a discovery. Two later measurements support it, including an ablation that changes only the
knock rule while holding the rest of the expert fixed. That change drops the expert's win rate
against our agents from $70$ to $15$ percent. And the reward study below shows learners that refuse to chase gin, even when we pay
three times more for it. The assumption came first. The evidence came after, from learners that
never trained against the expert.

\paragraph{Why the expert's own blindness matters.}
The expert reads two of the four planes the learner sees: its own hand and the
top of the discard pile. It never models the opponent's hand or the unseen deck. Its observation
is a strict subset of the learner's, and two things follow with no experiment at all. First, the
expert scores about $50$ percent against a copy of itself, so parity is a well-defined target.
Second, any policy reaching parity is a function of information the learner already receives, so
extra observations are not necessary for parity. Whether a learner of this size can be trained
to that policy is a separate question, and it stays open.

\subsection{Algorithm: trust-region updates win}
\begin{figure*}[t]
\centering
\includegraphics[width=0.49\textwidth]{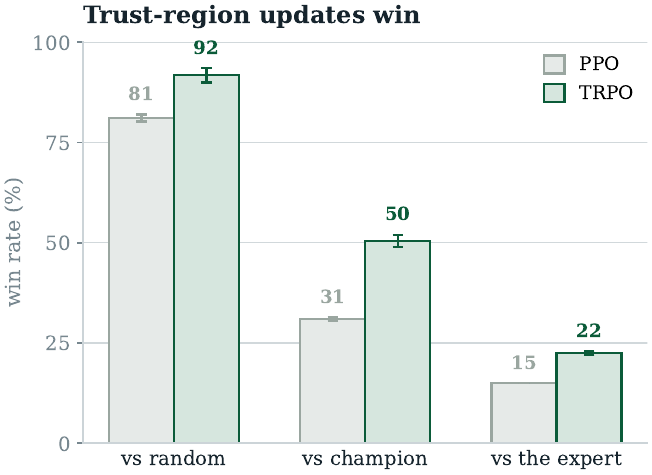}\hfill
\includegraphics[width=0.49\textwidth]{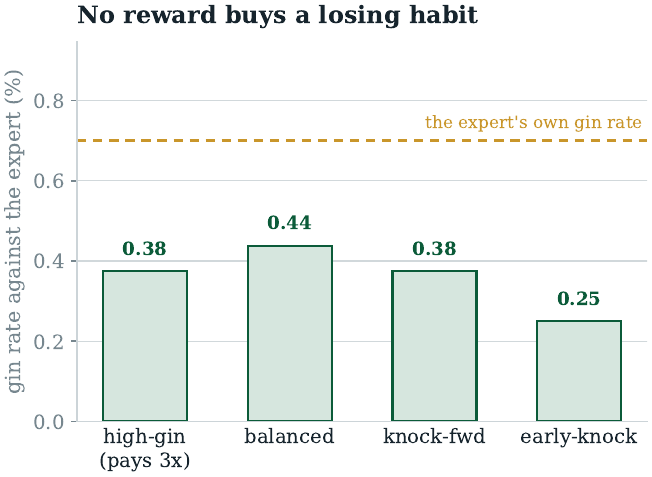}
\caption{Left: TRPO beats PPO against every opponent. Right: the gin rate stays
under one percent for every reward design we tried, including one that pays three
times more for a gin than a knock. The dashed line marks the expert's own gin
rate. You cannot pay the agent into chasing gin.}
\label{fig:algoreward}
\end{figure*}

With the same masking, curriculum and reward at two million steps, TRPO beats PPO
on every opponent and both seeds. Against the expert, TRPO reaches $22.5$ percent and PPO $15.0$ percent. Against
the champion, the rates are $50.5$ and $31.0$ percent (Figure~\ref{fig:algoreward}, left). Rewards here are rare and the opponent keeps changing. One
reading is that such a setting suits the smaller steps a trust region takes. 

\subsection{Reward: you cannot pay the agent into ginning}
\label{sec:reward}
The gin-to-knock payoff is a real dial for \emph{style}. Reward gin heavily and
the agent gins more often against weak opponents; reward knocking and it knocks almost always.
Against random play, a knock-first reward gives $97$ percent knocks at a $99.4$
percent win-rate. A gin-first reward gives $22$ percent gins at a $98.3$ percent win-rate. But style is not strength. Against the expert, the gin rate collapses under every
reward setting. Paying three times more for a gin than a knock leaves the gin rate under one
percent (Figure~\ref{fig:algoreward}, right). That is where it sits with no gin reward at all. The
learners land where the expert does: against strong play, chasing gin loses. A shaped reward can
set the agent's style, but it does not buy a losing habit.

\subsection{Representation: the raw sparse input wins}
Compressing the sparse $4\times52$ observation into a small dense embedding hurt
every time. Against the expert, the raw sparse input reaches about $15$ percent, while learned
and LLM-judged embeddings land between $6$ and $14$ percent depending on size. None beat the
sparse baseline, and neither larger embeddings nor unfreezing the encoder closed the gap. A fixed low-dimensional bottleneck can say whether two states are broadly
similar. But it discards which exact cards are where, and that is what the policy acts on. The observation is already
small, so there is little to compress.

\subsection{Curriculum, warm-start, and keep-the-best}
\begin{figure*}[t]
\centering
\includegraphics[width=0.78\textwidth]{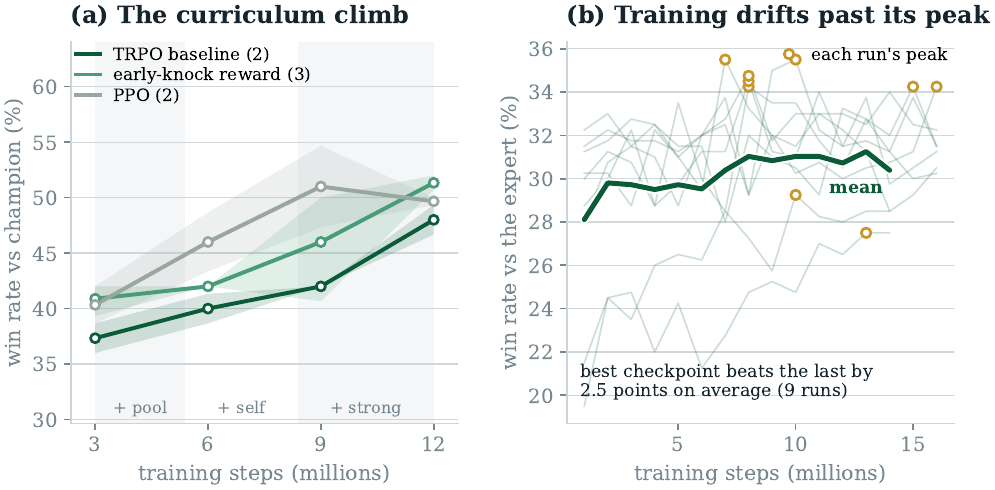}
\caption{(a) The curriculum climb: win-rate against the champion as training
moves through the stages, with the line the mean over seeds and the band their range.
(b) Why we ship the best checkpoint rather than the last. These runs evaluate every
million steps instead of every three, so the peak and the drift after it are visible.
Gold circles mark each run's best checkpoint.}
\label{fig:curves}
\end{figure*}

Self-play supplies opponents at no extra cost, and a three-million-step agent
beats its own parent $61$ percent of the time. Unguided self-play is fragile, though, and a pool
of past selves with no schedule broke down after about ten million steps. The rising curriculum moves from random play to a pool and then self-play,
keeping the challenge slightly ahead of the agent without breaking down. Warm-starting from
the champion gives each run a strong start, while keeping the best checkpoint protects against
drift past the peak. We evaluate every million steps and ship the best. Across nine
densely evaluated runs the best checkpoint beats the last by $2.5$ points on average, and by
as much as $5.2$ (Figure~\ref{fig:curves}b). PFSP
opponent-picking was about even with a simple schedule here, and a higher discount slightly
hurt.

We then combine every ingredient that helps. The final recipe uses TRPO, a
knock-first reward with a small deadwood bonus, a curriculum seeded with the strongest earlier
agents, warm-starting, and keep-the-best. Checked over $2000$ games, the strongest agent reaches
$\mathbf{34.2 \pm 2.1}$ percent against the expert and $51.4$ percent against the old champion
(Table~\ref{tab:numbers}). The old champion sat at roughly $30$ percent. The interval is
evaluation noise on this checkpoint, not spread across training seeds. Two sibling agents
trained with PFSP reach $34.0$ percent, inside that interval. Table~\ref{tab:verdicts} lists
every ingredient and its verdict.

\begin{table}[t]
\centering
\small
\caption{Win-rate against the fixed expert for the main milestones (higher is
better), with win-rate against the prior self-play champion for context. The
headline agent is evaluated over $2000$ games with a $95$ percent confidence
interval; the others over $400$ to $600$ games.}
\label{tab:numbers}
\begin{tabular}{@{}lcc@{}}
\toprule
\textbf{Regime} & \textbf{vs.\ expert} & \textbf{vs.\ champion} \\
\midrule
PPO baseline, 2M steps          & 15.0 & 31.0 \\
TRPO baseline, 2M steps         & 22.5 & 50.5 \\
Self-play champion (prior best) & $\sim$30 & (self) \\
Stacked recipe (this work)      & \textbf{34.2 $\pm$ 2.1} & \textbf{51.4} \\
\bottomrule
\end{tabular}
\end{table}

\begin{table*}[t]
\centering
\small
\caption{Ingredient verdicts across more than one hundred controlled runs,
measured by win-rate against the fixed expert; the ``why'' column gives the
observed mechanism.}
\label{tab:verdicts}
\begin{tabular}{@{}lll@{}}
\toprule
\textbf{Ingredient} & \textbf{Verdict} & \textbf{Why} \\
\midrule
Keep the best checkpoint        & \helps & training drifts past its peak; the best checkpoint beats the last by $2.5$ points on average \\
Warm-start from the champion    & \helps & start strong, then specialize, rather than relearn from scratch \\
TRPO over PPO                    & \helps & smaller, safer steps suit rare rewards and a shifting opponent \\
Reward knocking, not gin         & \helps & matches the expert's low-risk, knock-early style \\
Rising opponent curriculum      & \helps & always a fair but slightly harder challenge; avoids self-play collapse \\
Pay three times more for gin     & \nofx  & the agent refuses the losing habit at any payoff; gin rate stays under one percent \\
PFSP opponent picking           & \nofx  & about even with a simple schedule at this scale \\
Learned state embeddings        & \fails & a fixed low-dimensional bottleneck throws away detail the policy needs \\
Imitation learning              & \fails & clones an imperfect LLM guide's moves, not the reasoning behind them \\
Dense short-horizon rewards     & \fails & short-sighted: farms instant points, blind to winning \\
Live LLM opponent               & \fails & competent but $9$ to $27$ seconds per move, far too slow for RL-scale rollouts \\
\bottomrule
\end{tabular}
\end{table*}

\section{Architecture, Baselines, and a Second Game}
We have tuned the algorithm, reward, curriculum and checkpointing to the
mid-thirties, and two questions follow. Is the network the missing lever, and how do principled
baselines compare? Neither question resolves the ceiling. Comparing the baselines instead leads us to
test the hidden information directly.

\begin{figure*}[t]
\centering
\includegraphics[width=0.80\textwidth]{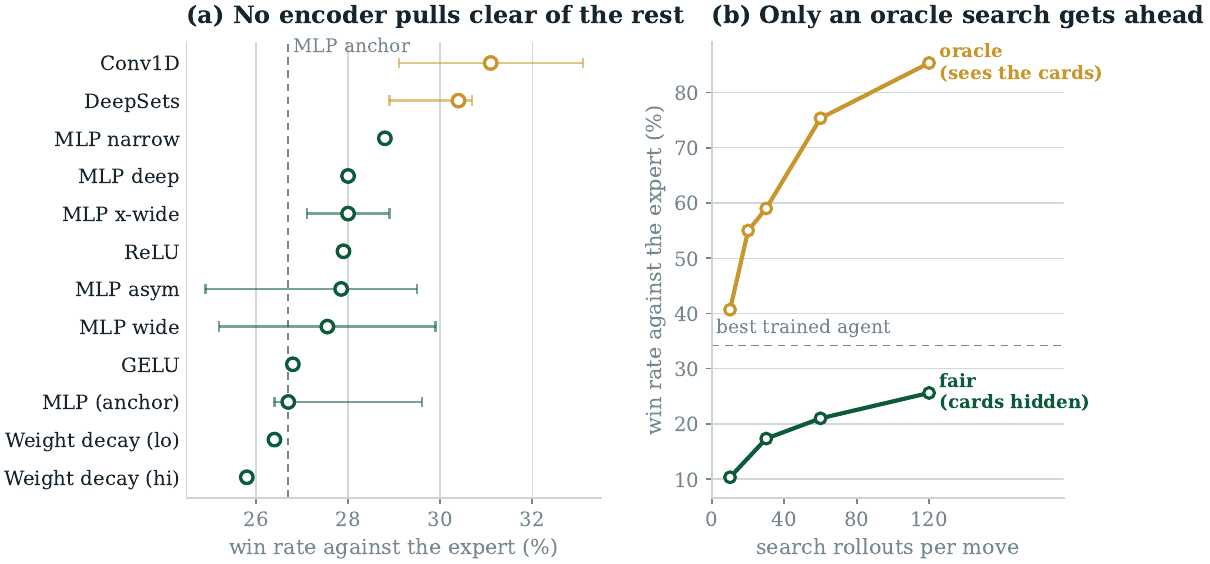}
\caption{(a) Win-rate against the fixed expert by architecture (IQM, $95$
percent stratified bootstrap CIs over seeds, from scratch at a fixed budget;
structured encoders in gold, MLP variants in green, anchor dashed). Every
encoder lands in one band with overlapping intervals. (b) A determinized search
graded \emph{fairly} (hidden cards re-dealt) tops out near $26$ percent, below
the trained agents (dashed), while the same search with \emph{oracle} access to
the hidden cards reaches $85$ percent. The gap between the curves is the value
of the hidden information.}
\label{fig:archbase}
\end{figure*}

\subsection{No architecture we could train breaks the ceiling}
We hold the winning recipe fixed and vary only the network, retraining from
scratch. The encoders are MLPs of several widths and depths, a convolutional stack, Deep Sets,
and an LSTM. Win-rate against the expert stays in a narrow band. The best interquartile means
are the structured encoders: convolutional at $31.1$ percent and Deep Sets at $30.4$ percent.
The plain MLP anchor is $26.7$ percent. But the $95$ percent bootstrap intervals overlap
throughout (Figure~\ref{fig:archbase}a), so we detect no difference between the architectures.
Wider, deeper, or extra-wide MLPs do not help, and weight decay slightly hurts. Rerun under PPO
or with PFSP sampling, the encoders keep their ordering, and the best cell reaches $30.2$
percent. That is the same band. Recurrence beats its matched feed-forward control, $24.3$
against $19.2$ percent, but both trail the TRPO anchor. A self-attention encoder cannot be
trained under our trust-region recipe at all. Differentiating the transformer twice for the
natural-gradient step is prohibitive on CPU and unstable on GPU. Graded under PPO against a matched MLP anchor at three million steps, it does
not separate from the MLP. Both reach a mean of $26$ percent over four seeds, with overlapping ranges, but
four seeds cannot establish equivalence. The data support only a failure to detect a
difference at this budget. No encoder we could train at
scale produced a detected gain over the plain MLP.

\paragraph{Is the ceiling an information limit? A direct test.}
\label{sec:oracleobs}
The oracle search prices hidden information only as an upper bound, since oracle
access also removes strategy fusion. So we put the question to the learner directly. We retrain
the identical PPO recipe with the opponent's hand appended as a fifth observation plane, and
change nothing else. We also trim each curriculum opponent's input to the width its own network
was trained on, so the extra plane reaches the learner alone. Handed the opponent's hand
outright, the agent reaches \ORACLEOBSMEAN{} percent against the expert over \ORACLEOBSN{}
seeds, against $25.5$ percent for the matched baseline. The oracle learner's range of $23.1$ to
$26.2$ sits inside the baseline's $23.5$ to $28.4$, so we detect no benefit from the extra
plane. We did check that plane against the engine's own hand at $1{,}200$ decision
points before training on it. We did not check that the network read it. Four seeds cannot prove the
absence of an effect either, and we saw nothing approaching the size the plateau would need.
Our conclusion is narrow: supplying the hidden hand does not lift this learner.
Within this stack, the plateau is therefore not explained by that information being absent
from the input.

\paragraph{Capacity buys wall-clock, not strength.}
The structured encoders also cost far more compute. On our CPU setup a full
ten-million-step run takes about four hours for the MLP. It takes sixteen for Deep Sets and
twenty-one for the convolutional stack. That is four to six times the compute for a win-rate that still overlaps the MLP
band. The same accounting prices the algorithm result. TRPO costs more
per step than PPO, because its update solves a conjugate-gradient system. Matching step budgets
therefore hands TRPO the larger share of wall-clock. Its win is a win at equal experience, not
at equal compute. Anyone budgeting in hours should read it that way. The efficiency carries to
play time. The shipped policy picks a move three to five orders of magnitude faster than the
searches and the LLM it is graded against (Table~\ref{tab:cost}).

\begin{table}[t]
\centering
\small
\caption{Seconds per move at play time. The policy is measured on one CPU
thread (median over $2000$ moves); the searches are derived from logged
evaluation time; the LLM is as measured in the Discussion. The shipped policy
fits $243$K parameters in a $1.9$ MB file.}
\label{tab:cost}
\begin{tabular}{@{}lc@{}}
\toprule
\textbf{Player} & \textbf{time per move} \\
\midrule
Our policy (MLP, best recipe) & $0.0003$ s \\
Fair ISMCTS, $60$ rollouts    & $1$ to $2$ s \\
Oracle ISMCTS, $120$ rollouts & $1.5$ to $3$ s \\
Live LLM opponent             & $9$ to $27$ s \\
\bottomrule
\end{tabular}
\end{table}

\subsection{A fair search is weak; only an oracle beats the expert}
We then run a determinized ISMCTS, the classic non-learned reference for these
games. We grade it \emph{fairly}, re-dealing the unseen cards before every rollout, and it is
weak. Win-rate against the expert rises as the per-move budget grows
(Figure~\ref{fig:archbase}b). At $10$, $30$, $60$ and $120$ rollouts, it reaches only $10$,
$17$, $21$ and $26$ percent. That is
\emph{below} our trained agents. This is the textbook weakness of determinized search. It
re-solves each sampled world on its own, so it assumes it could act differently in worlds it
cannot tell apart. That is the \emph{strategy-fusion} error \citep{strategyfusion}. Search over
public belief states, as in ReBeL \citep{rebel}, avoids it and is the natural stronger
baseline. Head to head, each
trained agent in Table~\ref{tab:baselines} beats the fair search ($66$ to $69$ percent at $60$
rollouts). For contrast, an oracle search that may see the hidden cards reaches $41$ to
$85$ percent over the same budgets. The $26$-versus-$85$ gap bounds what that information is
worth to a searcher able to act on it. It is only a bound, because oracle access also removes
the fusion error. It does say that naive search is not a free way past the ceiling.

\subsection{Generality: a second game with a computable optimum}
Finally we apply the grading protocol to Leduc Hold'em. There a
counterfactual-regret solution is computable, and our CFR expert reaches exploitability $0.026$.
This is a methodological check: it runs the same yardstick protocol with a
computed near-optimum rather than a heuristic reference. Graded against it over eight seeds, a tabular
self-play learner reaches near parity: mean return $-0.085$, where random is about $-0.78$. NFSP
averages $-0.71$ after three million episodes at its published defaults. We did not tune it for
this domain, so read it as an untuned baseline. Leduc is far smaller, so the learner gets closer
to its ceiling. What transfers is the protocol, not the size of the gap, and the factor sweep
was not re-run there (Table~\ref{tab:baselines}).

\begin{table}[t]
\centering
\small
\caption{Baselines and the second game. Top: trained agents vs.\ the \emph{fair}
ISMCTS (model win-rate, $60$ rollouts). Bottom: Leduc Hold'em mean return vs.\ the
CFR-optimal expert (exploitability $0.026$; $0$ is parity, random $\approx-0.78$).
Over seeds the tabular learner spans $+0.005$ to $-0.169$ and NFSP $-0.62$ to
$-0.78$.}
\label{tab:baselines}
\begin{tabular}{@{}lc@{}}
\toprule
\textbf{Trained agent vs.\ fair ISMCTS} & \textbf{win\%} \\
\midrule
Curriculum Ace (best)            & 69 \\
League Tactician (PFSP)          & 69 \\
Self-play champion               & 66 \\
\midrule
\textbf{Leduc Hold'em vs.\ CFR optimum} & \textbf{return} \\
\midrule
Tabular Q self-play (8 seeds)    & $-0.085$ \\
NFSP, 3M episodes (4 seeds)      & $-0.71$ \\
Random baseline                  & $-0.78$ \\
\bottomrule
\end{tabular}
\end{table}

\section{Discussion, Limitations, and Future Work}
\paragraph{What sets the ceiling, and what we cannot say.}
Stacking every choice that helps moves a small agent from roughly $30$ to about
$34$ percent against the expert (Table~\ref{tab:numbers}). What holds the rest of the gap is the
question we answer least well. Our two direct attempts on it point different ways. No encoder we
could train at scale separates from the others (Figure~\ref{fig:archbase}a). A fair determinized
search is weaker than our agents. Yet the same search reading the hidden cards reaches $85$
percent (Figure~\ref{fig:archbase}b). The hidden cards are valuable to a player able to act on them, but handing them to
the learner as an extra input changes nothing. Together these results describe our learner's
failure to convert valuable hidden information, not the value of the information itself. Capacity, optimisation, and the reactive form of
the policy all remain in play, and we cannot separate them here. Attention was not trainable
under our trust-region recipe, and only matched a plain MLP under PPO. Our recurrent result is
one LSTM configuration. The ceiling is a measurement, not a disappointment. It marks where a
small reactive agent lands under this recipe. Passing it takes something we did not test, rather
than more reward tuning.

\paragraph{What did not work, and why.}
\emph{Imitation learning} drove the behavior-cloning loss to near zero, yet the
resulting policy won almost no games. Its pure-RL baseline reaches only about $8$ percent
against the expert. The
teacher was a strong but imperfect LLM guide. The student may have cloned spurious correlates of
its actions rather than the reasoning behind them \citep{causalconfusion}. Or it may simply have
been too small to represent that reasoning. Our data do not separate the two. \emph{Dense step
rewards} made the agent short-sighted. A horizon of two never learned. A horizon of five stalled
after about five hundred thousand steps, with the agent farming instant points instead of
winning. That is what shaping theory warns of when the bonus is not aligned with the return
\citep{shaping}. A \emph{live LLM opponent} reached $98.2$ percent legal moves with a
chain-of-thought prompt, against $79.3$ with a terse one. One even beat our self-play agent
three games to two. But at $9$ to $27$ seconds per move it is far too slow for RL-scale rollouts. A
vision-language variant fed the board as an image managed about ten percent legal moves.

\paragraph{Limitations.}
The factor sweep is one game, so the specific numbers are about Gin Rummy. Our
expert is a strong heuristic, not a game-theoretic optimum, so win-rate against it is a fixed
yardstick, not a distance from perfect play. Our headline learner is reactive and feed-forward,
with no memory and no opponent model. And we used the LLM as a live opponent, an imitation
teacher, and a similarity judge, but never as a source of offline training data.

\paragraph{Future work.}
Naive search did not break the ceiling at the budgets we tried, and the hidden
cards lifted a searcher but not our learner. So the open question is what kind of agent converts
them. Candidates are explicit opponent-hand estimation, belief-state search as in ReBeL
\citep{rebel}, and regret-minimization methods \citep{cfr,deepcfr,nfsp}. Stronger memory is a second axis, and offline distillation is a third.
LLM-versus-LLM games could train a policy without the latency that killed the live opponent.

\paragraph{When no fixed yardstick exists.}
Not every game offers a cheap, strong reference, but the protocol needs a
\emph{fixed} one rather than a perfect one. The strongest available heuristic serves, and so does
a frozen Elo-ranked pool or a checkpoint frozen before the comparison. Fix it before tuning, so
the moving target of self-play cannot flatter the result.

\paragraph{Beyond one game.}
The protocol and the pipeline are not specific to Gin Rummy. Point them at
another two-player game in the same interface and the training and grading machinery is
untouched. Three things have to be rebuilt for each game, namely the observation planes, the
reward terms and the expert.

\section{Conclusion}
We asked two questions about a small, fast Gin Rummy agent. How strong can it
get? And which choices make it strong? A fixed expert and more than a hundred controlled runs
turned both into measurements. Strong play knocks early and almost never gins, and no reward
could tempt the agent into chasing gin. The recipe that works is trust-region self-play with a
knock-first reward, a rising curriculum, warm-starting, and keeping the best checkpoint. It
reaches about $34$ percent against the expert. Learned embeddings, imitation, dense rewards, and
a live LLM opponent each fail. What sets the remaining ceiling is harder to pin down than we
first thought. No network we could train moves it, and a fair search trails our agents. An oracle
holding the hidden cards pulls away, which bounds their value without pricing it. Yet handing
those cards to the learner produces no gain, so missing input information does not explain the
cap we measure. The grading protocol carries over
to Leduc Hold'em.

Strength in these games is easy to overstate against random or self-referential
opponents. A fixed, strong, reproducible reference turns a list of plausible tricks into a study
that says which choices matter. Build that reference first.



\end{document}